\documentclass[runningheads]{llncs}
\usepackage{graphicx}
\usepackage{amsmath,amssymb}
\usepackage{color}
\usepackage{times}
\usepackage{epsfig}
\usepackage{subfigure}
\usepackage{algorithm}
\usepackage{algpseudocode}
\usepackage{rotating}
\usepackage{multirow}
\usepackage{amsopn}
\usepackage{caption}

\begin{document}

\newcommand{\point}{
    \raise0.7ex\hbox{.}
    }

\pagestyle{headings}

\mainmatter

\title{Towards Segmenting Consumer Stereo Videos:\\Benchmark, Baselines and Ensembles} %
\titlerunning{Towards Segmenting Consumer Stereo Videos: Benchmark, Baselines and Ensembles}
\authorrunning{Wei-Chen Chiu, Fabio Galasso, Mario Fritz}

\author{Wei-Chen Chiu$^{\dag}$, Fabio Galasso$^{\ddag}$, Mario Fritz$^{\dag}$}
\institute{$^{\dag}$ Max Planck Institute for Informatics, Saarland Informatics Campus, Germany\\$^{\ddag}$ OSRAM Corporate Technology}

\maketitle

\begin{abstract}
   Are we ready to segment \emph{consumer stereo videos}? The amount of this data type is rapidly increasing and encompasses rich information of appearance, motion and depth cues.
   However, the segmentation of such data is still largely unexplored.
   First, we propose therefore a new benchmark: videos, annotations and metrics to measure progress on this emerging challenge.
   Second, we evaluate several state of the art segmentation methods and propose a novel ensemble method based on recent spectral theory. This combines existing image and video segmentation techniques in an \emph{efficient} scheme. %
   Finally, we propose and integrate into this model a novel regressor, learnt to optimize the stereo segmentation performance directly via a differentiable proxy. The regressor makes our segmentation ensemble \emph{adaptive} to each stereo video and outperforms the segmentations of the ensemble as well as a most recent RGB-D segmentation technique.
\end{abstract}
\newcommand{\myparagraph}[1]{\vspace{0pt}\noindent{\bf #1}}

\section{Introduction}

\begin{figure}[t]
\centering
\includegraphics[width=0.75\linewidth]{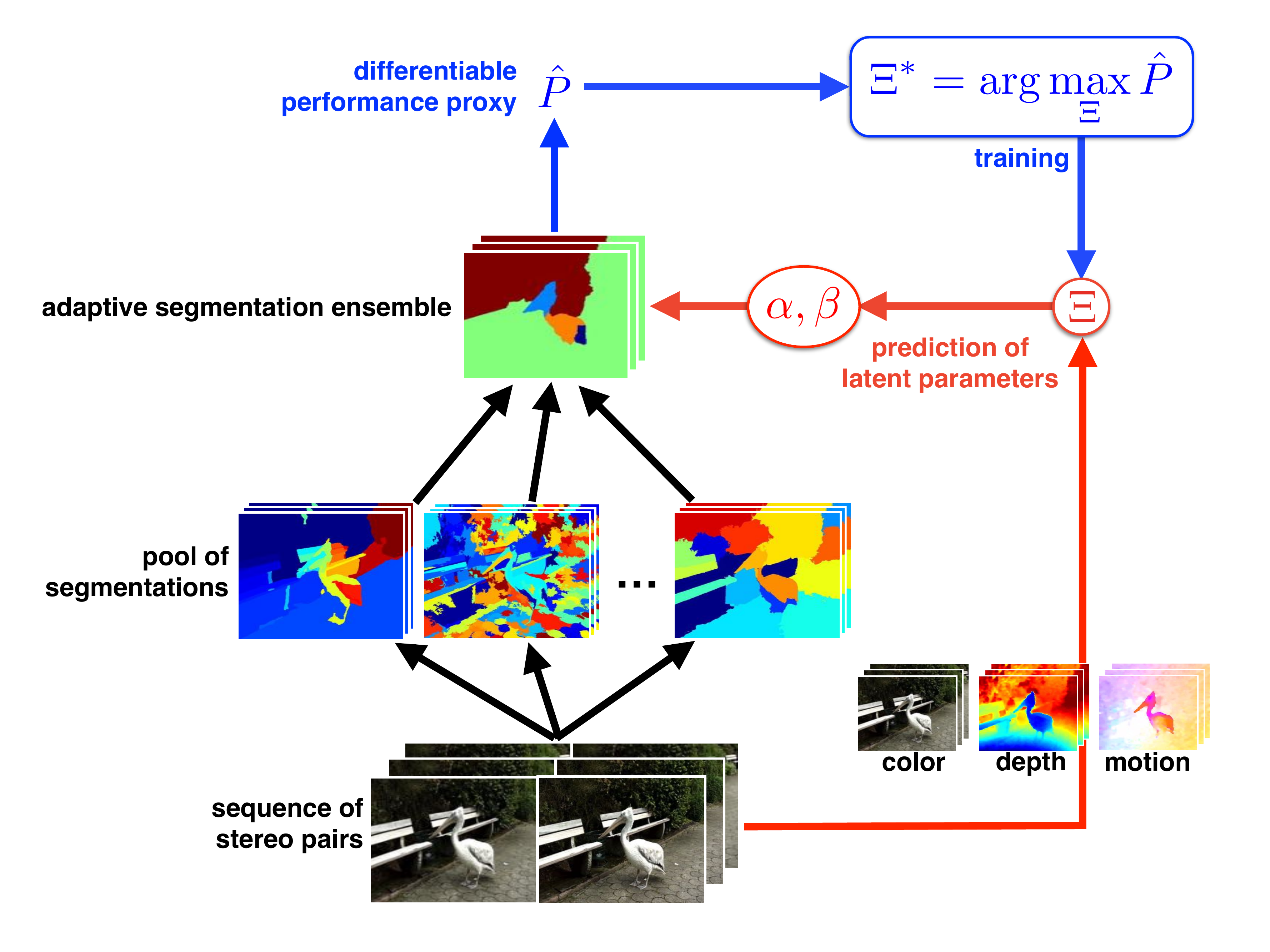}
\caption{Overview of the proposed efficient adaptive stereo segmentation technique. Our proposed segmentation ensemble model leverages the best available image and video segmentation results efficiently. A regressor makes the ensemble model adaptive to each stereo video, based on color, depth and motion features. In our novel learning framework, the segmentation performance is optimized  via a differentiable proxy.}
\label{fig:model_overview}
\end{figure}

We witness a fast growing number of stereo streams on the web, due to the advent of consumer stereo video cameras. Are we ready to expoit the rich cues which stereo videos deliver? Our work focuses on segmentation of such data sources, as it is a common pre-processing step for %
further analysis such as action~\cite{LeZYN11,OneataECCV14,TaralovaECCV14} or scene classification~\cite{RazaGrundmannEssaCVPR13}.

We propose a new consumer stereo video challenge, to understand the opportunities and foster the research in this new area. The new type of data combines the availability of appearance and motion with the possibility of extracting depth information. Considering consumer videos means addressing a most abundant web data, which is however also very heterogeneous, due to a variety of consumer cameras.

The new consumer stereo video challenge explicitly concerns the semantics of the video. A number of existing benchmarks have offered ground truth depth and motion, recurring to controlled recordings~\cite{ScharsteinGCPR14} or computer graphics simulations~\cite{Sintel}. By contrast, here we address stereo videos \emph{in the wild} and specifically consider the semantics of the data. While this might partly harm analysis (no true depth available), it addresses directly what we are most interested in, the actors and objects in the videos.

We warm-start the challenge with a number of baselines, extending best available image and video segmentations to the consumer stereo videos and their available features, e.g.\ color, motion and depth. Most baselines perform well on some videos, however none performs well on all. As an example, motion segmentation techniques~\cite{ochs2011object} perform well while the object moves, but encounter difficulty with static video shots. On the other hand, camouflaged (but moving) objects impinge appearance-based image~\cite{li2012segmentation} and video~\cite{grundmann2010efficient} segmentation techniques.

\begin{figure*}[t]
\begin{center}
\includegraphics[width=\textwidth,height=0.3\linewidth]{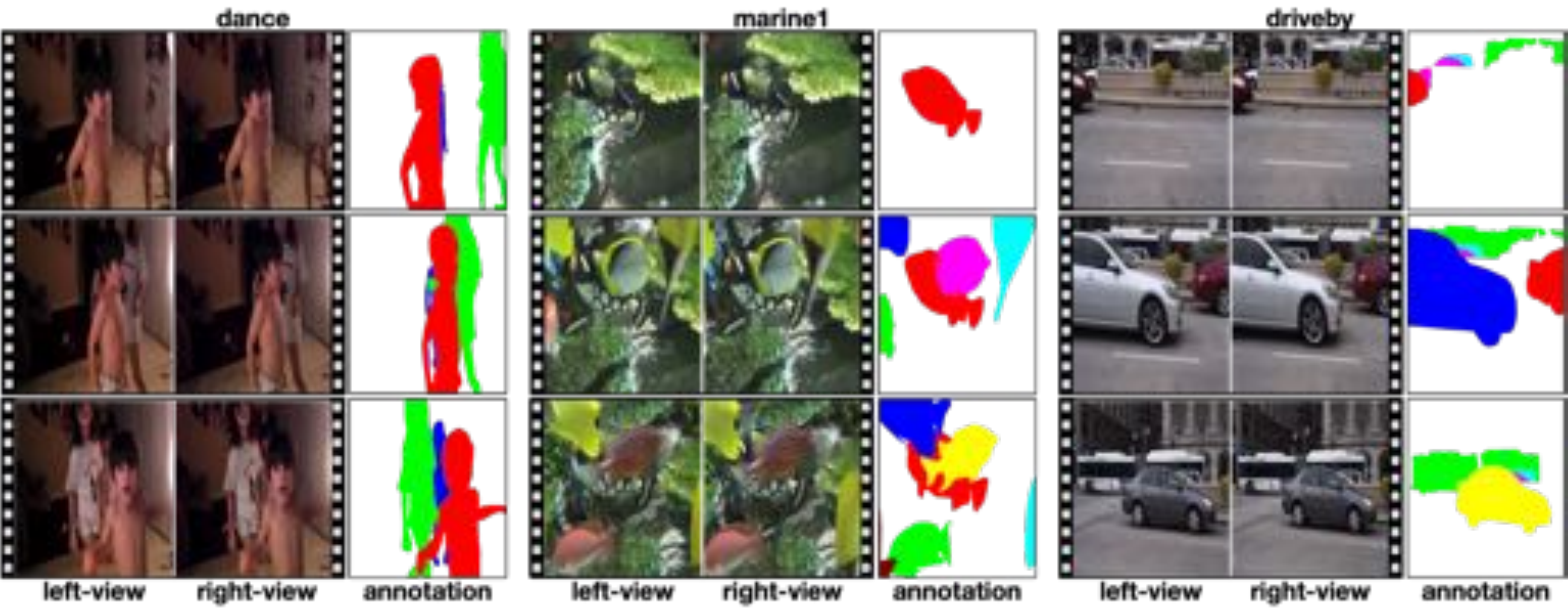}
\end{center}
\caption{Sample frames from the Consumer Stereo Video Segmentation Challenge (CSVSC) dataset (left-right views) with the corresponding annotations. The stereo videos differ in content (appearance, motion, number and type of subjects) and in camera characteristics (intrinsic parameters, zoom, noise).}\label{fig:dataset}
\end{figure*}

Thus motivated, we introduce in Section~\ref{sec:videosegm} a new efficient segmentation ensemble model, which leverages existing results where they perform best. Furthermore, we introduce in Section~\ref{sec:learningframework} the framework to learn a regressor which adapts the ensemble model to each particular stereo video. The proposed technique is overviewed in Section~\ref{sec:overviewpar} and demonstrated in Section~\ref{sec:expres}. Although \emph{only} combining optimally existing results, our new algorithm outperforms a most recent RGB-D segmentation technique~\cite{2014-Hickson-EHGSRV}.

\section{Related Work}

\noindent{\bf Video Segmentation}
Video segmentation has recently received much attention. It strikes how diversely the segmentation problem is defined.
\cite{brox2010object} looks at the problem of motion segmentation by using optical-flow long term trajectories. \cite{galasso2011spatio} also uses trajectories (defined densely with superpixel-regions) and looks at motion but focuses on people motion. \cite{LiICCV13} considers appearance to generate several image proposals and tracks the most temporally consistent ones with motion. \cite{grundmann2010efficient} addresses general unsupervised video segmentation based on appearance and motion. \cite{wang2015saliency} introduce an unsupervised, geodesic distance based, salient video object segmentation method. \cite{faktor2014video} propose a non-local consensus voting scheme defined over a graph of similar regions in the video sequence.
We note that: 1.\ the strong diversity of existing algorithms hardly allows combining their results; 2.\ none of those techniques may seamlessly generalize to stereo videos. This work proposes a solution to both aspects.\\

\noindent{\bf Depth/Stereo Segmentation}
There is a long tradition of work on 3D reconstruction which estimates 3D coordinates, thus depth, from pair or multiple views~\cite{kanade1994stereo,scharstein2002taxonomy}. These efforts have been recently combined with reasoning on the object appearance and the physical constraints of the 3D scene in the work of \cite{bleyer2012extracting}, whereby segmentation proposals are produced for semantic objects. The underlying assumption of a static scene for these methods does not allow their extension to stereo video sequences which we consider here. Additionally, the use of consumer cameras contrasts the high quality images which they generally require.\\

\noindent{\bf Scene Flow and Stereo Videos}
Recent work addresses scene flow, the joint estimation of optical flow and depth, assuming calibrated~\cite{huguet2007variational,basha2013multi} or uncalibrated cameras~\cite{vogel20113d}. Those do not address segmentation. Elsewhere, video segmentation is addressed by considering RGB-D information~\cite{bergh12wacv,WeikersdorferICIP13,2014-Hickson-EHGSRV,Held-RSS-16}, Kinect colour images with depth.
The stereo videos which we consider are not assumed calibrated nor from the same camera. Since we consider consumer cameras, the videos are further unconstrained in terms of spatio-temporal resolution, zooming, sensitivity and dynamical range, and present challenging motion blur effects and image latency.\\

\noindent{\bf Image/Video Co-Segmentation}
Recent researches on image and video co-segmentation \cite{gunheecvpr12,joulin12cvpr,fu2014object,chiu2013coseg} tasks provide a way to jointly extract common objects across multiple images or videos. The general assumption of co-segmentation problem on commonality of objects in a video set makes it a plausible fit to stereo video segmentation task when we consider left and right videos of stereo pairs as two separate sequences in the same set. However, from this perspective the depth cue in the stereo videos is not explicitly explored which can provide rich information to outline objects while other features are with ambiguities.\\

\noindent{\bf Algorithm Selection and Combination}
There has been previous work which attempted to select the best algorithm from a candidate pool depending on the specific task e.g. for recognition on a budget \cite{karayev12nips,karayev14cvpr} or active learning \cite{ebert12cvpr}. For optical flow, \cite{mac2010segmenting} presented a supervised learning approach to predict the most suitable algorithm, based on the confidence measures of the optical flow estimates. By contrast, Bai et al. \cite{bai2009video} shows an object cutout system where the segmentation is done by a set of local classifiers, each adaptively \emph{combining} multiple local features. Our proposed technique not only \emph{combines} different feature cues but also the available segmentation algorithms, weighting their contributions rather than selecting one. We show experimentally that combination provides better results. Our proposed approach relates closely to \cite{li2012segmentation} which efficiently combines image segmentation algorithms within the spectral clustering framework outperforming the original results in the pool. Our extension of~\cite{li2012segmentation} is twofold: 1.\ we generalize from image to stereo video segmentation; 2. we design and learn discriminatively an \emph{adaptive scheme} which allows to combine optimally the pool of video and segmentation algorithms \emph{at each superpixel}, based on the appearance and motion features of the local pixels. Experimentally, the adaptive scheme further improves on the static combination.

\section{Consumer Stereo Video Segmentation Challenge (CSVSC)} \label{sec:CSVS}

We launch a Consumer Stereo Video Segmentation  Challenge (CSVSC). The new dataset consists of 30 video sequences which we have selected from Youtube based on their heterogeneity. In fact, the footage differs in the number of objects (2-15), the kind of portrayed actors (animals or people), the type of motion (a few challenging stop-and-move scenes and objects entering or exiting the scene), the appearance visual complexity (also in relation to the background, a few objects may be harder to discern) and the distance of the objects from the camera (varying disparity and thus depth). Not less importantly, we have selected videos acquired by different consumer stereo cameras, which implies diverse camera intrinsic parameters, zooms and (as a further challenge) noise degradations such as motion blurs and camera shake. We illustrate a few sample sequences in Figure~\ref{fig:dataset}. %

\subsubsection*{Benchmark annotation and metrics.}\label{sec:metricsall}

We have gathered human annotations and defined metrics to quantify progress on the new benchmark. In particular, we equidistantly sample $5$ frames from the left view of all $30$ videos to be labelled ($150$ frames are labelled in whole benchmark, while stereo videos altogether total to $1738$ left-right-pair frames).

As for the metrics, we have considered state-of-the-art image~\cite{amfm_pami2011} and video segmentation~\cite{galasso13iccv} metrics:

\myparagraph{\textbf{Boundary precision-recall} (\textbf{BPR}).} This reflects the per-frame boundary alignment between a video segmentation solution and the human annotations. In particular, BPR indicates the F-measure between recall and precision~\cite{amfm_pami2011}.

\myparagraph{\textbf{Volume precision-recall} (\textbf{VPR}).} This measures the video segmentation property of temporal consistency. As for BPR, VPR also indicates the F-measure between recall and precision~\cite{galasso13iccv}.

It is of research interest to determine which of the metrics is best for learning. We answer this question in Section~\ref{sec:expres}, where we consider BPR and VPR alone or combined by their: \textbf{arithmetic mean} (\textbf{AM-BVPR}) or \textbf{harmonic mean} (\textbf{HM-BVPR}), formulated respectively as $\frac{\text{BPR}+\text{VPR}}{2}$ and $\frac{2\cdot \text{BPR}\cdot \text{VPR}}{\text{BPR}+\text{VPR}}$ .

\subsubsection*{Preparation of stereo videos}

\begin{figure}[t]
\begin{center}
\includegraphics[width=0.8\linewidth,height=0.25\linewidth]{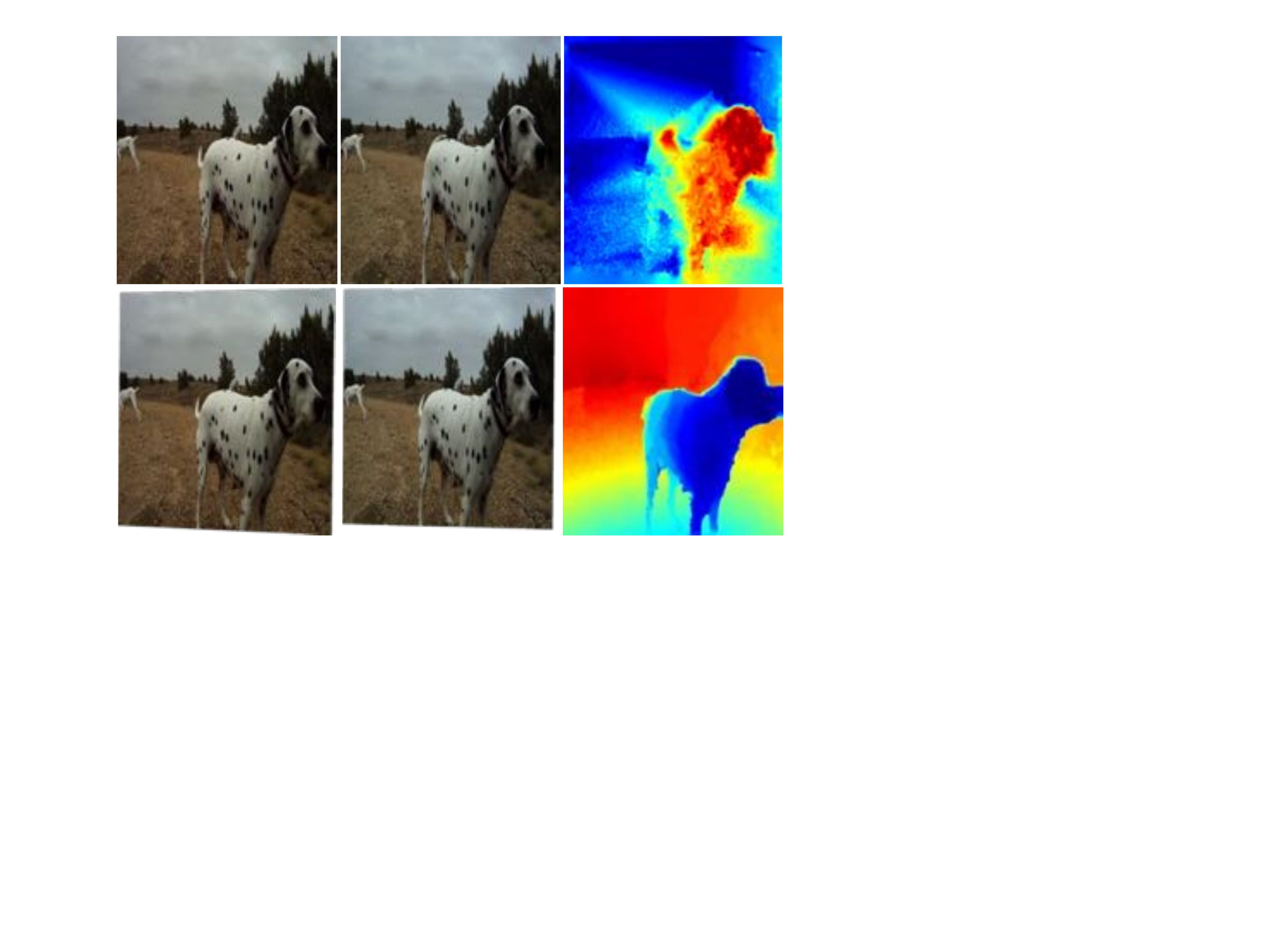}
\end{center}
\caption{Sample disparity estimation. The first two columns are the original stereo pair and their rectified images. The top-right picture is the disparity map computed by \cite{Geiger10}, the bottom-right is the depth map obtained by optical flow~\cite{zach2007duality} between the left and right view.}
\label{fig:depth}
\end{figure}

Not having a ground truth depth may impinge comparison among techniques applied to the dataset. We define therefore an initial set of comparisons among depth-estimation algorithms and make the results available.

We have considered the per-frame rectification of \cite{fusiello2008quasi} and the stereo matching algorithm of \cite{Geiger10}, filling-in the missing correspondences with \cite{2011JanochA-Cat}. Furthermore, we have estimated depth by the optical flow algorithm of \cite{zach2007duality} between the right and left views. We illustrate samples in Figure \ref{fig:depth}.
Our initial findings are that estimating depth by optical flow leads to best downstream stereo segmentation outputs, which we use therefore in the rest of the paper.

\section{Efficient adaptive segmentation of stereo videos}\label{sec:overviewpar}

We warm start the CSVSC challenge with a basic segmentation ensemble model. To this purpose we first pre-process the stereo videos with a pool of state-of-the-art image and video segmentation algorithms. Then we combine the segmentation outputs with a new efficient segmentation ensemble model (cf. Section~\ref{sec:videosegm}). Finally, we propose the learning framework to adapt the combination parameters of each stereo video (cf. Section~\ref{sec:learningframework}).

Figure~\ref{fig:model_overview} gives an overview of our ensemble model:

\myparagraph{Pool of image and video segmentations.} We select most recent algorithms which are available online. These are used to segment the single frames (image segmentations) and the left views of the stereo videos (video segmentations). This results in a pool of segments which are respectively superpixels and supervoxels.

\myparagraph{Efficient segmentation ensemble model.} We bring together the pool of segments and connect them to the stereo video voxels. The segmentation ensemble model is represented by a graph and parameterized by $\alpha$ and $\beta$'s, which weight the contribution of each segmentation method. The model is accurate (voxel-based) but costly. We propose therefore an efficient graph reduction which is exact, i.e.\ it provides the same solutions as the voxel-based at a lower computational complexity.

\myparagraph{Performance-driven adaptive combination.}
We compute stereo video features from the stereo videos based on color, flow and depth.
From these features, we regress the combination parameters $\alpha$ and $\beta$'s, i.e.\ we combine optimally the pooled segmentation outputs. To this purpose, we propose a novel regressor $\Xi$ and an inference procedure, to learn the optimal regression parameters $\xi$ from data. For the first time in literature, the regressor parameters $\xi$ are directly optimized according to the final performance measure $P$ (resulting from the graph partitioning and the metric evaluation, cf.\ Section~\ref{sec:learning}). We achieve this with a novel differentiable performance proxy $\hat P$.

None of the state-of-the-art segmentation algorithms performs well with all of the challenging consumer stereo videos (cf.\ experiments in Section \ref{sec:expres}). Both the contributions on the ensemble model (Section~\ref{sec:videosegm}) and the performance-driven adaptive combination (Section~\ref{sec:learningframework}) turn out important for better results.

\section{Efficient segmentation ensemble model}\label{sec:videosegm}

\begin{figure}[t]
\begin{center}
\includegraphics[width=0.75\columnwidth]{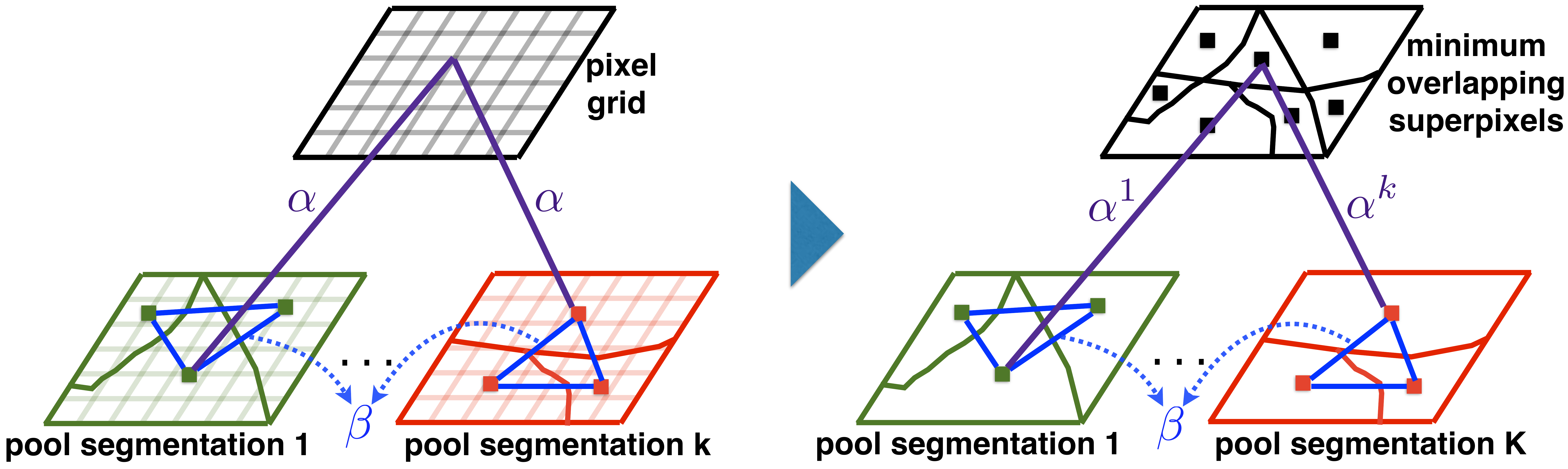}
\end{center}
\caption{Proposed video segmentation model. A number of $K$ pooled image and video segmentation outputs are brought together as hypotheses of grouping for the considered video sequence (cf.\ Section~\ref{sec:videosegm} for details). We propose to replace the model of \cite{li2012segmentation} (\textit{left}) with a new one (\textit{right}) based on minimally overlapping superpixels, which is provably equivalent but yields better efficiency (cf.\ Section~\ref{sec:unifyinggraph})}\label{fig:visualize_extension}
\end{figure}

We propose a graph for bringing together the available video segmentation outputs. Additionally, we propose the use of recent spectral techniques to reduce the voxel graph to one based on tailored superpixels/supervoxels, to improve efficiency without any (proven) compromise on performance. The graph partitioning with spectral clustering provides the segmentation output.

\subsection{Unifying graph}\label{sec:unifyinggraph}

Given a number of video segmentation outputs, we propose to bring all of them together by defining a unifying graph.

Let us consider Figure~\ref{fig:visualize_extension} \textit{left}. Each video segmentation algorithm provides groupings of the video sequence voxels. In the unifying graph, each pixel is therefore linked to the groupings to which it belongs. For example, one algorithm may compute spatio-temporal tubes (supervoxels)~\cite{grundmann2010efficient}, another one may compute image-based superpixels~\cite{amfm_pami2011}. The video sequence voxels would then be linked to the tube to which they belong (temporally) and to their superpixels (spatially).
Altogether, the outputs from the pool of video segmentations provide hypotheses of grouping for the video voxels.

More formally, we define a graph $\mathcal{G}=(\mathcal{V}, \mathcal{E})$ to jointly represent the video and the segmentation outputs.
Nodes from the vertex set $\mathcal{V}$ are of two kinds:

\myparagraph{Voxels} are the video sequence elements which we aim to segment;

\myparagraph{Pooled segmentation outputs} are the computed spatial- and/or temporal-groupings, providing voxel grouping hypotheses.

Further to being connected to the voxels, the pooled groupings from the same output are also connected to their neighbors, which defines the video volume structure. Edges are therefore of two types:

\myparagraph{$\beta$-edges} are between the groupings of each pooled segmentation $k$; we assume $C$ features (appearance, motion, etc.\ cf.\ Section~\ref{fig:visualize_extension}) and distances based on $\beta^c$-weighted features: $w^k_{I,J} = e^{-(\beta^1 d^{k_1}_{I,J} + \cdots + \beta^C d^{k_C}_{I,J})}$, where $d^{k_c}_{I,J}$ is the distance between superpixels $I$ and $J$ from the $k$-th pooled output based on $c$-th feature.

\myparagraph{$\alpha$-edges} are between the voxels and the grouping that it belongs to. The $\alpha^k$'s encode the trust towards the respective $K$ segmentation algorithm, ideally proportional to its accuracy. In other words, the weights of $alpha$ edges correspond to the importance of individual segmentations, thus a higher $\alpha^k$ can be interpreted as a larger contribution to the overall performance, and shows the importance of individual segmentations

Partitioning graph $\mathcal{G}$ with spectral clustering is computationally demanding as the number of nodes (and edges) depends linearly on the video voxels. The theory of \cite{li2012segmentation} reduces the complexity of a first stage of spectral clustering (the eigendecomposition) but not of the second one (k-means), still of linear complexity in the number of voxels (and thus bottleneck of \cite{li2012segmentation}). We address both with graph reduction in the following Section.

\subsection{Improved efficiency with graph reduction}\label{sec:graphreduction}

Let us consider again Figure~\ref{fig:visualize_extension}. A huge number of voxels are similar both in appearance and in motion and are therefore grouped in all segmentation outputs. When partitioning the original graph $\mathcal{G}$, these voxels are always segmented together. (The trivial proof leverages their equal edges and therefore eigenvectors).

Rather than considering all voxels, we propose to reduce the original graph $\mathcal{G}$ to one of smaller size $\mathcal{G}^Q$ which is equivalent (provides exactly the same clustering solutions). In $\mathcal{G}^Q$, we basically group all those voxels with equal connections into super-nodes (reweighting their edges equivalently). This reduces the algorithmic complexity, as the spectral clustering (both the eigendecomposition and the k-means) now depends only on the number of super-nodes (which is determined for most pooled segmentation algorithms by the number of objects, rather than voxels).

We identify the voxels with equal connections by intersecting the available segmentation outputs. The result of the intersection is an oversegmentation into superpixels, which can generate all pooled segmentation outputs by merging. We name these \emph{minimally overlapping superpixels}.

More formally, the reduced graph $\mathcal{G}^Q=(\mathcal{V}^Q, \mathcal{E}^Q)$ takes the minimally overlapping superpixels as nodes $\mathcal{V}^Q$ and the following edge weights
\begin{equation}
\scalebox{0.9}{$
w^Q_{IJ}=\left\{
  \begin{array}{l l}
   \displaystyle\sum_{i\in I}\sum_{j\in J}w_{ij} & \quad \text{if} \,I\neq J\\
   \displaystyle\frac{1}{|I|}\sum_{i\in I}\sum_{j\in J}{w_{ij}} - \frac{(|I|-1)}{|I|} \sum_{i\in I}\sum_{j\in \mathcal{V}\setminus I}w_{ij}& \quad \text{if}\, I=J
  \end{array} \right.
$}
\label{eq:graph_reduction}
\end{equation}
where $\left | \cdot \right |$ indicates the number of pixels within the superpixel, $I, J$ are two minimally overlapping superpixels, and $w_{ij}$ stands for a generic edge of the original graph. According to \eqref{eq:graph_reduction}, two pixels $i$ and $j$ are reduced if belonging to the same superpixel, i.e.\ if $I=J$. These self-edges are of great importance, because spectral clustering normalizes clusters by their accumulated volumes of merged pixels, i.e. summations of merged $\alpha$'s. (Since the superpixel connections are equal for the pixels within the same superpixel by construction, the reduction is exact, cf.~\cite{GalassoetalCVPR14}.)

\subsection{Implementation details}\label{sec:implement_details_sec4}

The output segmentation is obtained by graph partitioning $\mathcal{G}^Q$ with spectral clustering~\cite{ng2002spectral,ShiMalikTPAMI2000,LuxburgStatComp07}. In particular, the labels of the minimally overlapping superpixels provide the voxel labels and thus the video segmentation solution.

In this work, we use $K=6$ image and video segmentation algorithms: ($1.$) The hierarchical image segmentation of \cite{amfm_pami2011}. We choose one layer from hierarchy based on best performance on a validation set. We take three segmentation outputs by applying the Simple Linear Iterative Clustering (SLIC)~\cite{achanta2010slic} respectively on ($2.$) depth, ($3.$) optical-flow~\cite{zach2007duality} and the ($4.$) LAB-color coded cues, bilaterally filtered for noise removal and edge preservation~\cite{zhang2014rolling}. ($5.$) Hierarchical graph-based video segmentation (GBH)~\cite{grundmann2010efficient}. We choose one layer from the hierarchy on the validation set. ($6.$) The motion segmentation technique (moseg) of \cite{ochs2011object}.

While the features are computed on the stereo video. The graph is constructed on one of the two views (the left one) of the stereo videos, which is then evaluated for the segmentation quality. The contribution of segmentation outputs is weighted by $\alpha$. $\beta$ defines the affinities between superpixels/supervoxels from the same pooled segmentation output, weighting $C=3$ feature cues based on mean Lab-color, depth and motion.

Note the importance of $\alpha$'s and $\beta$'s in the graph $\mathcal{G}$ and therefore $\mathcal{G}^Q$. These parameters define how much each pooled segmentation output is trusted and how to compute the similarity among superpixels/supervoxels in these outputs. Such parameters can be defined statically (cf.~\cite{li2012segmentation}) or adjusted dynamically in a data-dependent fashion, as we propose in the next Section.

\section{Performance-driven adaptive combination}\label{sec:learningframework}

We propose a regressor $\Xi$ to estimate the optimal segmentation ensemble parameters $\alpha$ and $\beta$ from the appearance-, motion- and depth-based features of the stereo videos. (Cf.\ Figure~\ref{fig:model_overview} where the regressor is given by the red arrows.) Furthermore, we propose a novel inference framework to learn the regressor parameters $\xi$ from the training stereo videos. (Cf.\ Figure~\ref{fig:model_overview} where the training is represented with blue arrows.) A new differentiable performance proxy $\hat P$ enables optimization driven by the stereo video segmentation performance measure $P$.

\subsection{Adaptive combination by regression}\label{sec:regressor}

Let us define a regressor $\Xi$, with parameters $\xi$. $\Xi$ takes as input a set of features $\mathcal{F}$ computed from the stereo video and outputs the parameters $\alpha$ and $\beta$ for the ensemble segmentation model (i.e.\ the coefficients to optimally combine $K$ segmentation outputs from the pool based on $C$ features, cf.\ \ref{sec:unifyinggraph}).
Intuitively, the regressor should select the best segmentation outputs from the pool, based on the stereo video content. This would imply, for example, a larger trust towards image- rather than motion-segmentation outputs, for those stereo videos where no motion is present.

While any type of regressor could be adopted, here in our approach we employ a second order regressor $\Xi$ which we parameterize by a matrix $B$. Overall, $\alpha$ and $\beta$ are computed as:
\begin{equation}
(\alpha^1, \dots, \alpha^K, \beta^1, \dots, \beta^C) = \Xi (\mathcal{F}; \xi) =  \mathcal{F}^T \mathit{B}  \mathcal{F}
\end{equation}
where $\mathit{B}$ is learnt with least squares and the input-output pairs of $\mathcal{F}$ and $(\alpha,\beta)$, and $\xi$ are the regression coefficients contained in $\mathit{B}$.
We consider in $\mathcal{F}$ features based on appearance, motion and depth. A large feature set is important to allow the regressor to understand the type of stereo video (dynamic, static, textured etc.) For each feature, we compute therefore histograms, means, medians, variances and entropies. We would leave the learning framework to choose from the right feature, i.e.\ training the best regressor $\Xi$. This should ideally consider the system performance $P$ for optimization or the tractable differentiable proxy which we discuss next.

\subsection{Performance-driven regressor learning by differentiable proxies} \label{sec:learning}

Let us consider Figure~\ref{fig:model_overview}. The $\alpha$ and $\beta$, regressed by $\Xi$ according to features $\mathcal{F}$, correspond to a stereo video segmentation performance $P$. During \emph{training}, we seek to optimize $\Xi$ for the maximum segmentation performance $P$:
\begin{equation}
\hat \Xi = \max_\Xi P(\Xi(\mathcal{F}))
\label{eq:regressor}
\end{equation}

There are two main obstacles to our goal.
First, typical video segmentation performance metrics are not differentiable and therefore do not lend themselves to directly optimizing an overall performance. To address this, we propose a differentiable performance proxy $\hat P$ in Section~\ref{sec:performance_proxy}.

Second, $\alpha$ and $\beta$ are not part of the objective \eqref{eq:regressor} and have to be considered \emph{latent}. In Section~\ref{sec:jointgloballearn}, we define therefore an EM-based strategy to jointly learn the regressor $\Xi$, $\alpha$ and $\beta$.
An overview of our training procedure is given in Algorithm~\ref{algo:joint_learning}.
\begin{algorithm}[h]
  \begin{algorithmic}[1]
  	\Require $\forall$ training videos with initial set of parameters $(\alpha, \beta)$ and stereo video features $\mathcal{F}$
  	\Repeat
  	\State Given the current estimates of $(\alpha, \beta)$, train the $\Xi$ which regresses them from $\mathcal{F}$
    \ForAll {training video}
    	\State predict $({\alpha}'', {\beta}'') = \Xi(\mathcal{F}$);
        \State use $({\alpha}'', {\beta}'')$ as initialization for $(\hat{\alpha}, \hat{\beta}) = \arg\max_{\alpha,\beta} \hat P(\alpha,\beta)$;
        \State update $(\alpha, \beta)$ for the training video by $(\hat{\alpha}, \hat{\beta})$;
    \EndFor
    \Until{Convergence or max. iterations exceeded}
    \caption{Joint learning of the regressor $\Xi$ and latent combination weights $\alpha,\beta$}
	\label{algo:joint_learning}
  \end{algorithmic}
\end{algorithm}

\subsubsection{Metric Specific Performance Proxy}\label{sec:performance_proxy}

In image segmentation, performance is generally measured by boundary precision recall (BPR) and its associated best F-measure~\cite{amfm_pami2011}. In video segmentation, benchmarks additionally include volume precision recall (VPR) metrics~\cite{galasso13iccv}. Both these performance measures are plausible $P$, but neither of them is differentiable, which complicates optimization. (We experiment on various performance measures in Section~\ref{sec:expres}.)

We propose to estimate a differentiable performance proxy $\hat P$ which approximates the true performance $P$. We do so by a second order approximation parameterized by the matrix $Y$. Taking $\chi$ a vector of features which are sufficient to represent the stereo video (at least as far as the estimation of $(\alpha, \beta)$ is concerned, will be described in the next paragraph) we have:
\begin{eqnarray}
\label{eq:local_regressor}
(\hat \alpha, \hat \beta) = \arg\max_{\alpha, \beta} \hat P(\alpha,\beta) = \arg\max_{\alpha, \beta}\chi^\top Y\chi
\end{eqnarray}
We perform training by sampling $\alpha$ and $\beta$, computing then input-output pairs of vector $\chi$ and the real performance values $P$, and finally fitting the parameter matrix $Y$.

\paragraph{Stereo video representation by spectral properties.}

We are motivated by prior work on supervised learning in spectral clustering~\cite{meilia2005regularized,jordan2004learning,IonescuICCV2015} to represent the stereo videos by their spectral properties.
In particular, we draw on \cite{meilia2005regularized} and consider the normalized-cut cost $\textrm{NCut}$ (of the similarity graph, based on the training set groundtruth labelling) and its lower bound $\textrm{Trace}_R$. Our representation vector is therefore $\chi = \left [ \alpha, \beta, \textrm{NCut}, \textrm{Trace}_R\right ]^\top$.

In more details, given the indicator matrix $E=\left \{ e_r \right \}_{r = 1 \cdots R}$ where $e_r \in \mathbb{R}^{N^m}$, $e_r(i)=1$ if superpixel $i$ belongs to $r$-th cluster otherwise $=0$, and $N^m$ is the number of superpixels, we have:
\begin{align}
\begin{split}
\textrm{NCut}(\alpha, \beta, E) &= \sum_{r=1}^R \frac{e_r^\top(D-W)e_r}{e_r^\top De_r}\\
 \textrm{Trace}_R(\alpha, \beta) &=  R - \sum_{r=1}^R \lambda_r(L)
\end{split}
\end{align}
where $D = \mathbf{diag}(W\mathbf{1})$ is the degree matrix of $W$ and $\lambda_r(L)$ is the $r$-th eigenvalues of the generalized Laplacian matrix $L = D^{-1} \cdot W$ of the similarity matrix $W$.

\paragraph{Derivatives of Performance Proxy.}

Our performance proxy $\hat P$ is now differentiable. For gradient descent optimization, we use its derivatives w.r.t.\ parameters $\theta \in \left \{ \alpha^k, \beta^c \right \}$:
\begin{equation}
\frac{\partial \chi^\top Y\chi}{\partial \theta} =  \frac{\partial \chi^\top}{\partial \theta}(Y+Y^\top)\chi \quad \forall \theta \in \left \{ \alpha^k, \beta^c \right \}
\end{equation}
The derivatives of $\textrm{NCut}$ and $\textrm{Trace}_R$ $\in \chi$ are:
\begin{align}
\begin{split}
\frac{\partial (\textrm{NCut})}{\partial \theta} &= \sum_{r=1}^R \frac{-e_r^\top \frac{\partial W}{\partial \theta}e_re_r^\top De_r + e_r^\top We_re_r^\top \frac{\partial D}{\partial \theta}e_r} {(e_r^\top De_r)^2}\\
\frac{\partial (\textrm{Trace}_R)}{\partial \theta} &= \text{ trace}(V^\top \frac{\partial L(\theta)}{\partial \theta}V)
\label{eq:diff_objs}
\end{split}
\end{align}
where $V$ denotes the subspace spanned by the first $R$ eigenvectors of $L$.
Note that $W$, $D$ and $L$ are all parameterized by $\alpha$ and $\beta$. By plain chain rule differentiation, we compute the gradients of the differentiable proxy $\hat P$ in closed form. (Cf.\ All gradients are presented in the supplementary material.)

\subsubsection{Joint Learning of Regressor and Latent Parameter Combinations}\label{sec:jointgloballearn}

As stated in Equation~\ref{eq:regressor}, we are interested in optimizing the performance $P$ w.r.t.\ the regressor $\Xi$ and therefore the ensemble combination parameters $\alpha$ and $\beta$ have to be treated as latent variables. As described in Algorithm~\ref{algo:joint_learning}, we solve this by an EM-type optimization scheme in which we iterate finding optimal parameters $\alpha$ and $\beta$ and predicting new $\alpha$ and $\beta$ parameters based on the re-fitted regressor $\Xi$.

Intuitively, this scheme strikes a balance between the generalization capabilities of the regressor and optimal parameters $\alpha$ and $\beta$. We found this to be particular important, as in many cases a wide range of parameters leads to good results. Fixing the best parameters as a learning target, leads to a more difficult regression and overall worse performance.
The metric specific performance proxy is continuously updated by using the samples in a small neighborhood in order to improve the local approximation of the desired metric $P$.

\subsection{Implementation Details} \label{sec:implement}

As already noted, the computation of $\textrm{NCut}$ at training involves the ground truth annotations. In particular, the $\textrm{NCut}$ for the entire video requires all frames labeled, while ours and most segmentation datasets~\cite{galasso13iccv,OchsTPAMI2014} only offer sparse labeling. Aggregating dense optical flow over time allows to connect the sparsely annotated frames. The spatial and temporal connections of these labeled frames are then used for the $\textrm{NCut}$ computation.

Our representation vector $\chi$ in \eqref{eq:local_regressor} consists of $\left [ \alpha, \beta, \textrm{NCut}, \textrm{Trace}_R\right ]$. We have empirically found that this combination improves of the individual parts and subsets by $5\%$ and therefore we use the full vector in the following experiments.

In order to increase the number of examples for our training procedure, we divide each video into subsequences so that each of them contains two frames with groundtruth.

\section{Experimental Results} \label{sec:expres}

\begin{figure*}[t]
\centering
\centerline{\includegraphics[width=\linewidth]{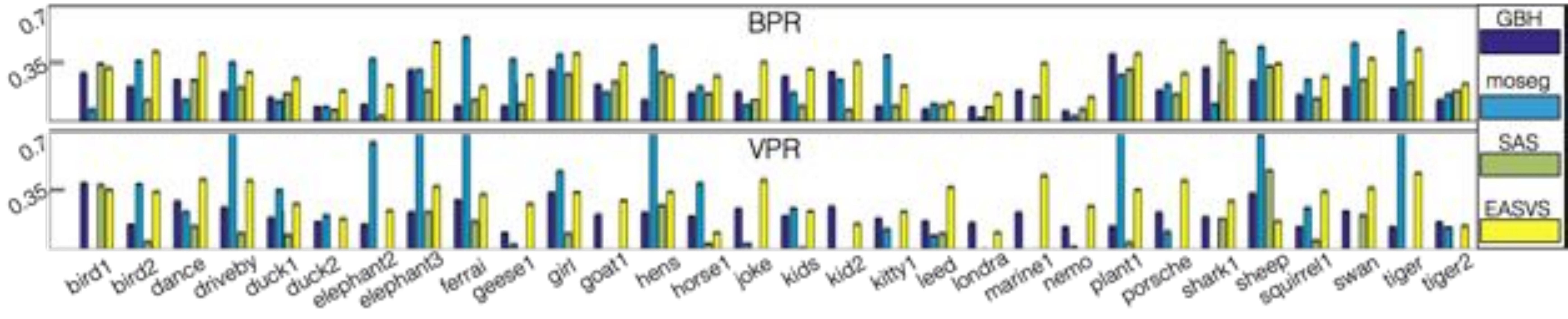}}
\caption{Results of the considered video segmentation algorithms (GBH~\cite{grundmann2010efficient}, moseg~\cite{ochs2011object}, SAS~\cite{li2012segmentation}) and our proposed EASVS on the CSVSC stereo video sequences using BPR and VPR. No considered method performs consistently well on all videos. moseg may achieve high performance of stereo videos with large and distinctive motion such as "elephants3" and "hens" but underperforms when motion is not strong, e.g.\ "marine1". Complementary features are given by GBH. SAS combines statically (cf.\ segmentation ensemble model of Section~\ref{sec:unifyinggraph}) the two video segmentation techniques as well as the pooled image segments but also underperforms, because a static combination cannot address the variety of the stereo videos.
}
\label{fig:existing}
\end{figure*}

We evaluate our proposed \textbf{efficient and adaptive stereo video segmentation} algorithm (\textbf{EASVS}) on the CSVSC benchmark. In particular, first we test the pooled segmentation outputs, then we compare EASVS against relevant state-of-the-art on stereo video sequences, finally we present an in-depth analysis of EASVS.

\subsection{Video segmentations and their (static) ensemble}\label{sec:eval_video_seg}

Among the pooled segmentations (details in Section~\ref{sec:implement_details_sec4}), we have included two state-of-the-art video segmentation techniques: the motion segmentation algorithm of \cite{ochs2011object} (moseg) and the graph-based hierarchical video segmentation method of \cite{grundmann2010efficient} (GBH).

In Figure~\ref{fig:existing}, we illustrate performance of each of moseg and GBH on all stereo video sequences. (Cf.\ detailed comments in the figure caption.) As expected, none of the two performs satisfactorily on all sequences. Rather, they have in most cases complementary performance, e.g. moseg aiming for motion segmentation takes the lead on sequences with evident motion and good optical flow estimates; while GBH overtakes when spatio-temporal appearance cues are more peculiar in the visual objects.

A third technique illustrated in Figure~\ref{fig:existing} is the segmentation by aggregating superpixel method of \cite{li2012segmentation} (SAS). This is an interesting baseline for our proposed algorithm. SAS is based on a static combination of pooled segmentation outputs. We extend its original image-based formulation to stereo videos by including into its pool the GBH and moseg video segmentation methods, as we illustrate in Section~\ref{sec:unifyinggraph}.

Figure~\ref{fig:existing} clearly states that a static combination does not suffice to address the segmentation of stereo videos. By contrast, quite surprisingly, trying to always pool \emph{all} video and image segmentation output \emph{with the same contributing weights} turns out to harm performance.

\begin{table}[h]
\centering
\begin{tabular}{|c|c|c|c|c|}
\hline
stereo video segmentation   & BPR   & VPR   & AM-BVPR & HM-BVPR \\ \hline
GBH \cite{grundmann2010efficient} & 0.187 & 0.208 & 0.198 & 0.198 \\ \hline
moseg \cite{ochs2011object} & 0.247 & 0.285 & 0.266 & 0.264 \\ \hline
SAS \cite{li2012segmentation} & 0.184 & 0.087 & 0.135 & 0.118 \\ \hline \hline
4D-seg \cite{2014-Hickson-EHGSRV} & 0.128 & 0.146 & 0.137 & 0.120 \\ \hline
VideoCoSeg \cite{chiu2013coseg} & 0.238 & 0.140 & 0.189 & 0.169 \\ \hline \hline
Proposed EASVS  & \textbf{0.301} & \textbf{0.296} & \textbf{0.295} & \textbf{0.288} \\ \hline \hline
Oracle   & 0.371 & 0.505 & 0.423 & 0.428 \\ \hline
\end{tabular}
\caption{Results on the CSVSC benchmark. The proposed EASVS outperforms the baselines from video segmentation, static ensemble, RGB-D video segmentation, and video co-segmentation for all metrics. See detailed discussion in Section~\ref{sec:sotacomp}.}
\label{tab:vid_results}
\end{table}

\begin{table}[h]
\centering
\begin{tabular}{|c|c|c|c|c|c|c|}
\hline
 & no depth & fixed depth & fixed $\alpha$ & fixed $\beta$ & fixed both $\alpha, \beta$ & proposed EASVS \\ \hline
HM-BVPR  & 0.254 & 0.276 & 0.270 & 0.276 & 0.272 & \textbf{0.288}\\ \hline
\end{tabular}
\caption{Analysis of the proposed EASVS, which shows the importance of the depth cue as well as the proposed adaptive strategy. See Section~\ref{sec:indepth} for detailed discussion.}
\label{tab:adapt_results}
\end{table}

\begin{table*}[ht]
\centering
\setlength{\tabcolsep}{1pt}
\begin{tabular}{cc|cccc|cccc}
\includegraphics[width=0.095\linewidth]{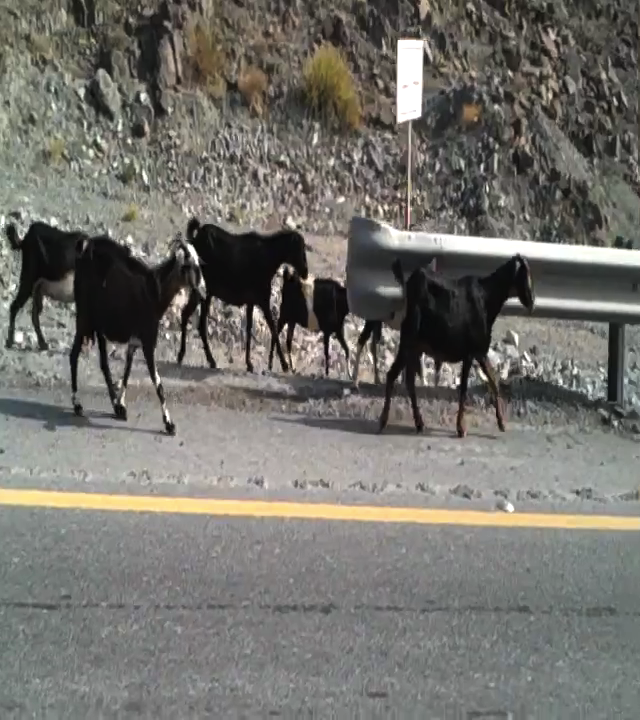} &
\includegraphics[width=0.095\linewidth]{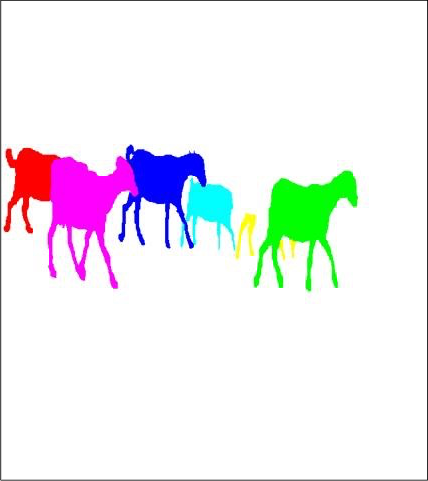} &
\includegraphics[width=0.095\linewidth]{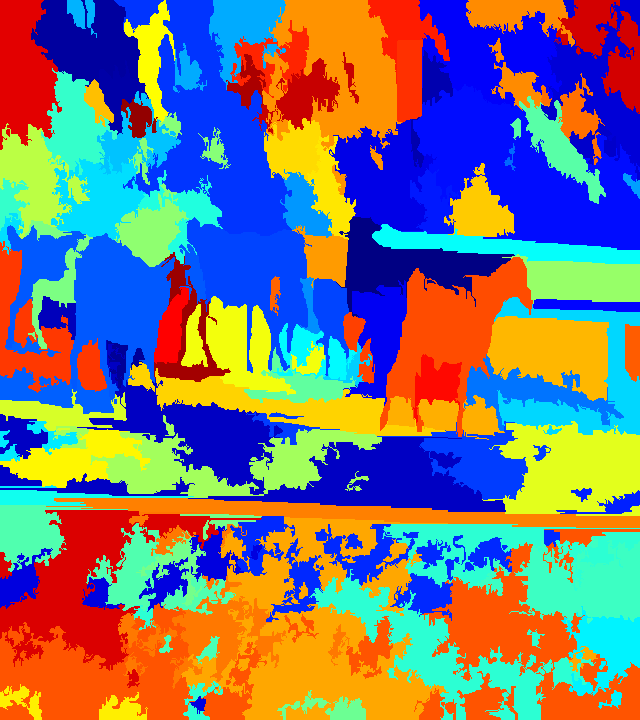} &
\includegraphics[width=0.095\linewidth]{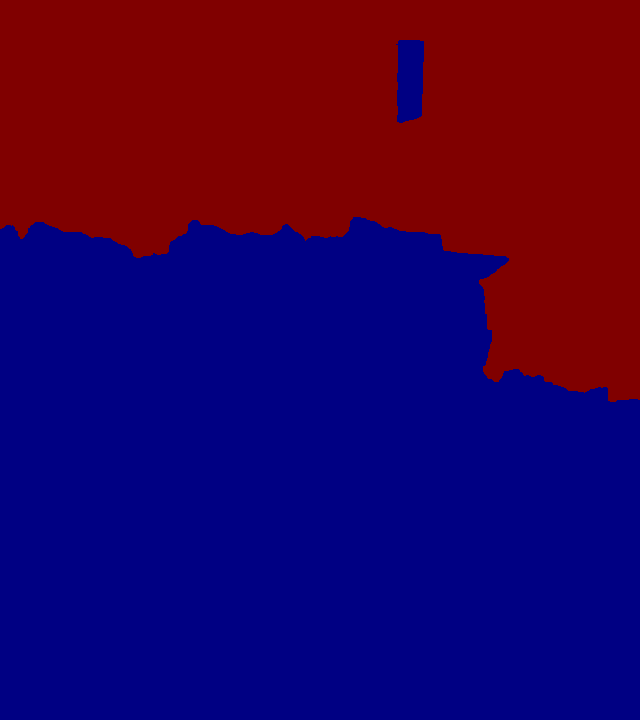} &
\includegraphics[width=0.095\linewidth]{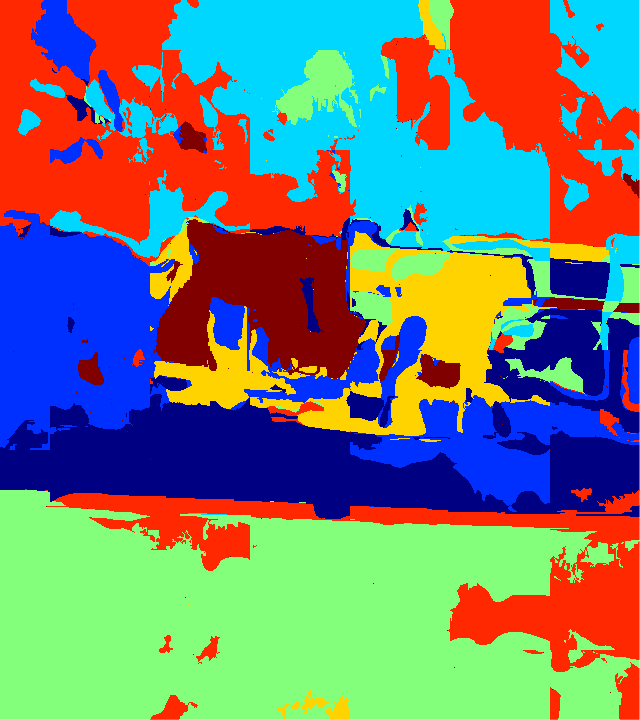} &
\includegraphics[width=0.095\linewidth]{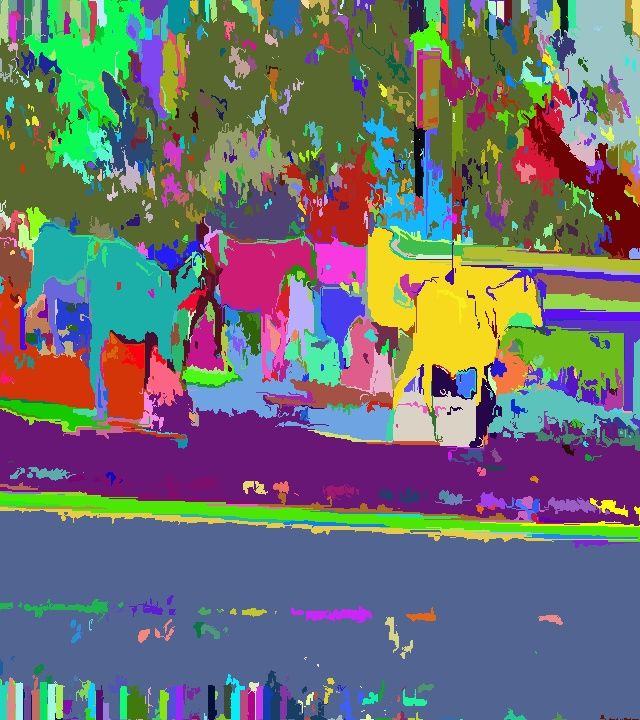} &
\includegraphics[width=0.095\linewidth]{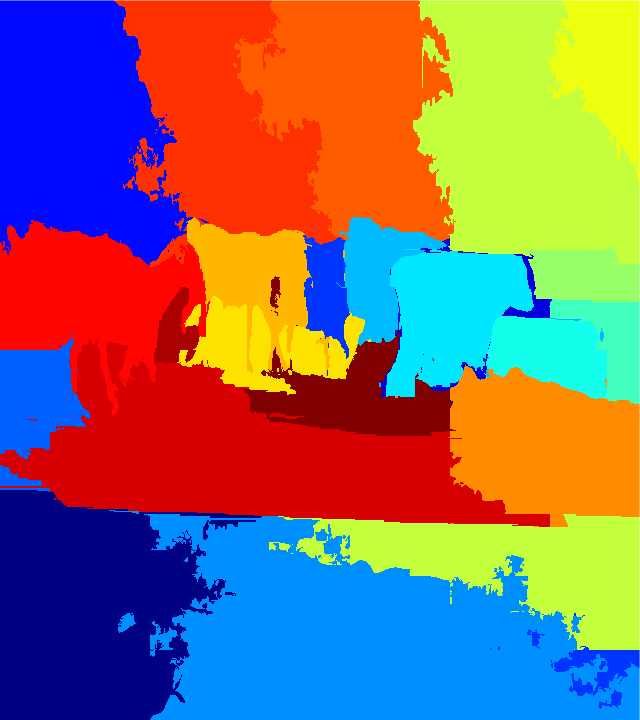} &
\includegraphics[width=0.095\linewidth]{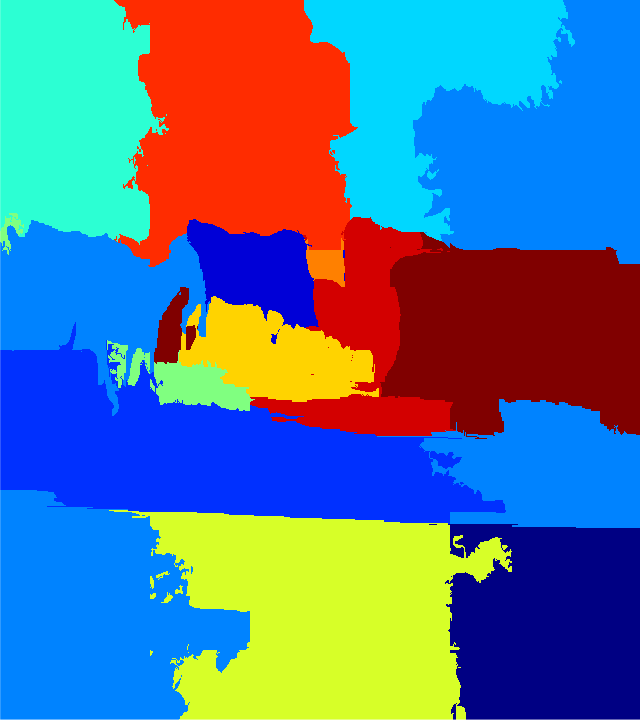} &
\includegraphics[width=0.095\linewidth]{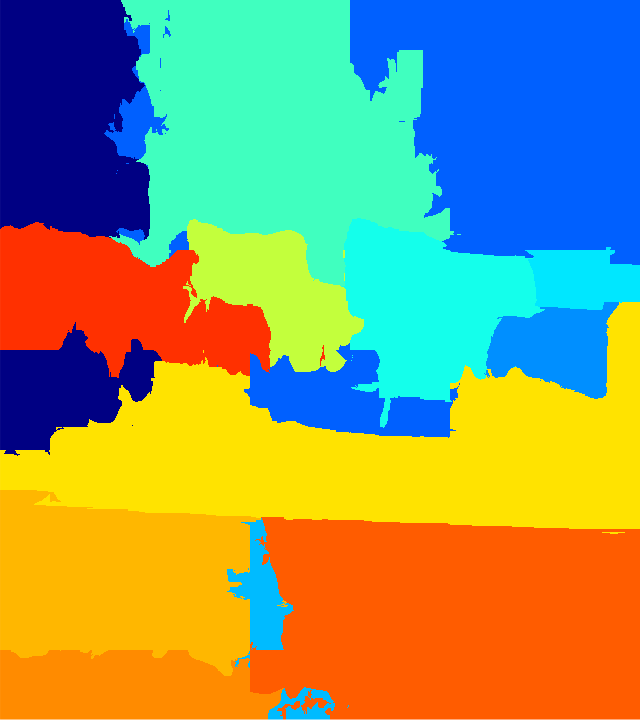} &
\includegraphics[width=0.095\linewidth]{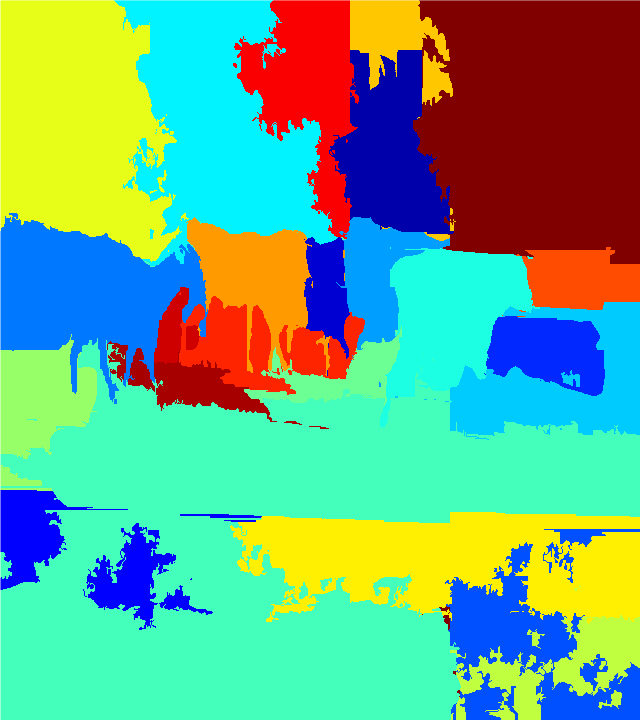}
\\
\includegraphics[width=0.095\linewidth]{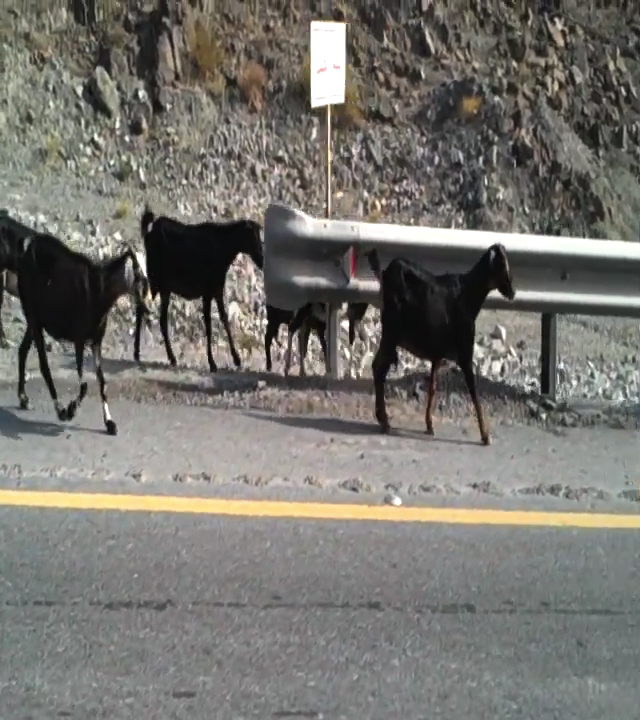} &
\includegraphics[width=0.095\linewidth]{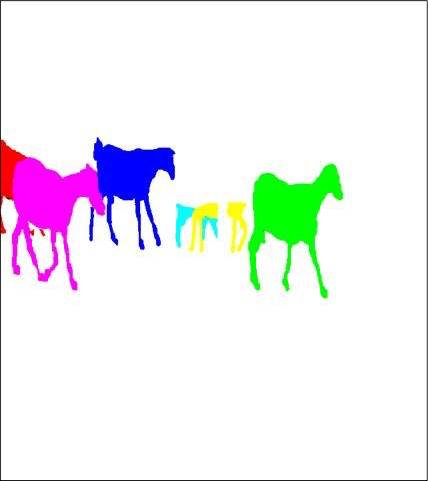} &
\includegraphics[width=0.095\linewidth]{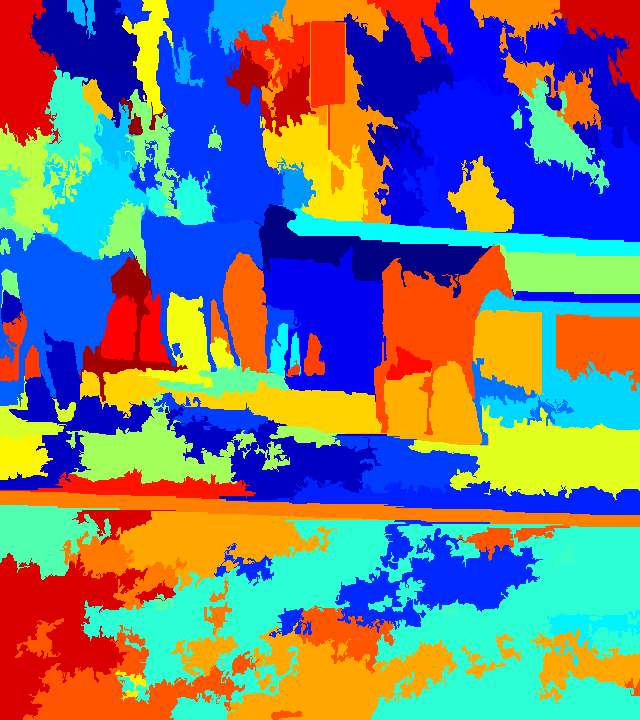} &
\includegraphics[width=0.095\linewidth]{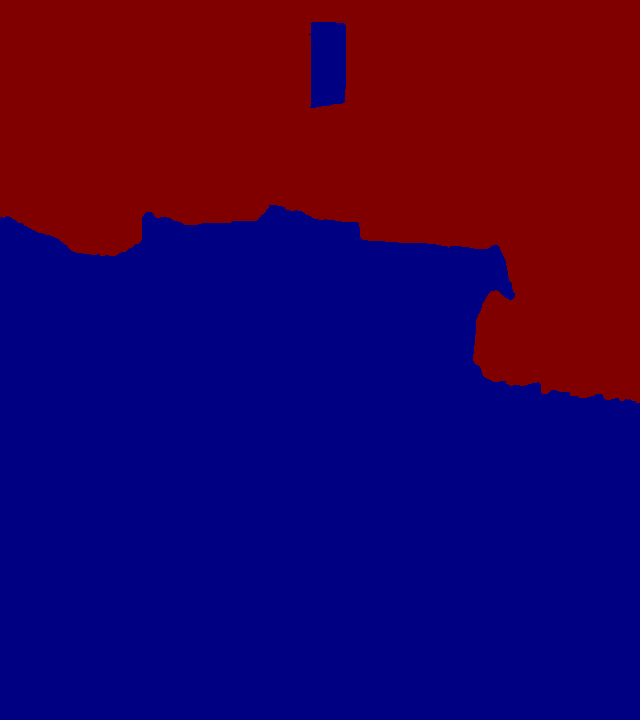} &
\includegraphics[width=0.095\linewidth]{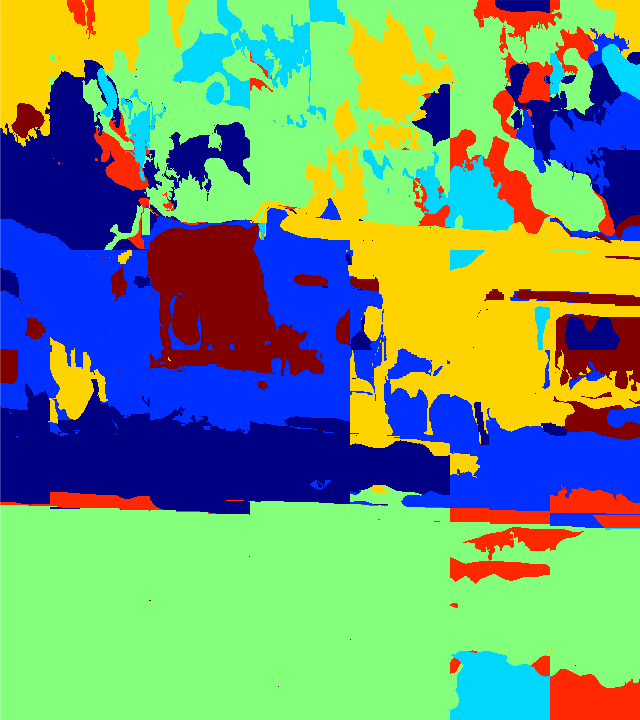} &
\includegraphics[width=0.095\linewidth]{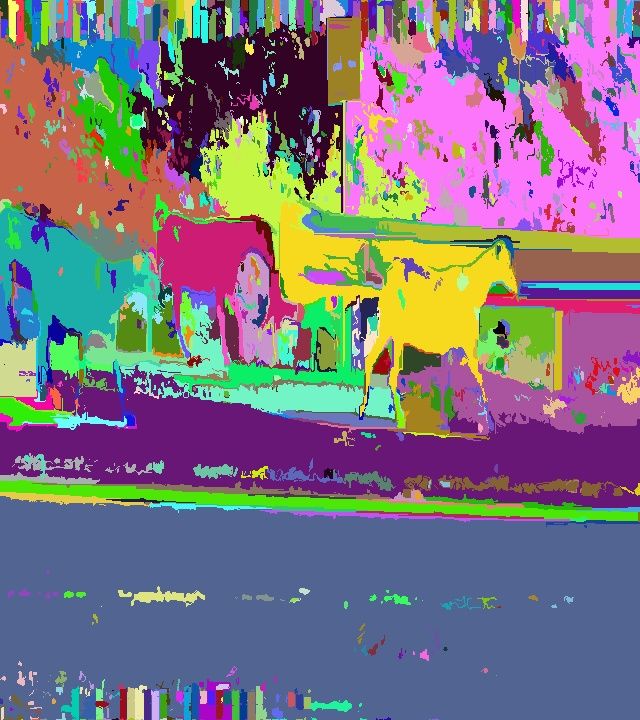} &
\includegraphics[width=0.095\linewidth]{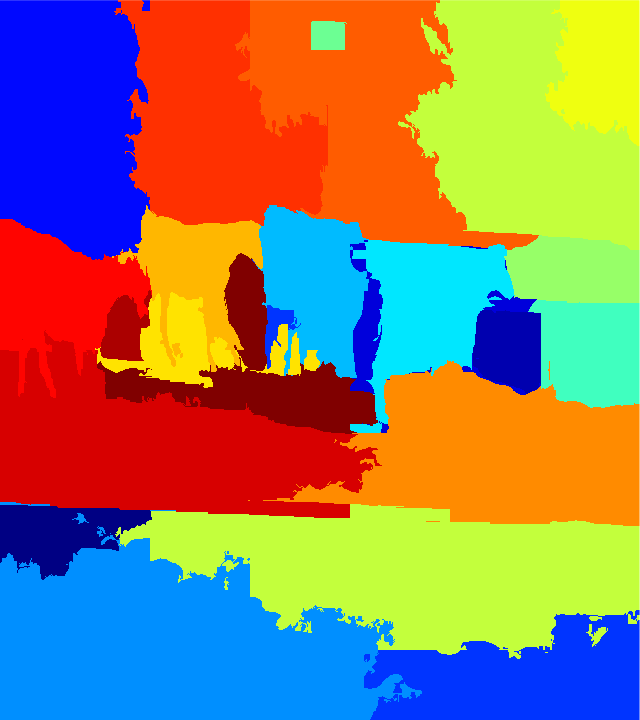} &
\includegraphics[width=0.095\linewidth]{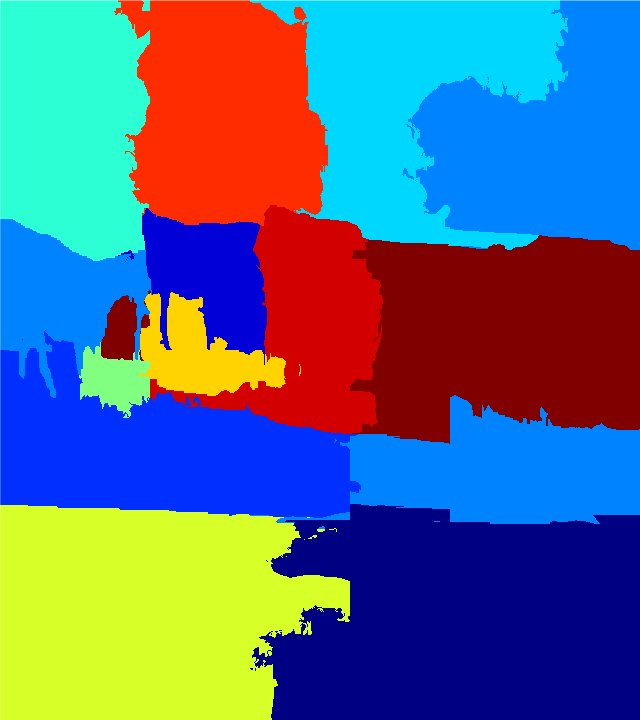} &
\includegraphics[width=0.095\linewidth]{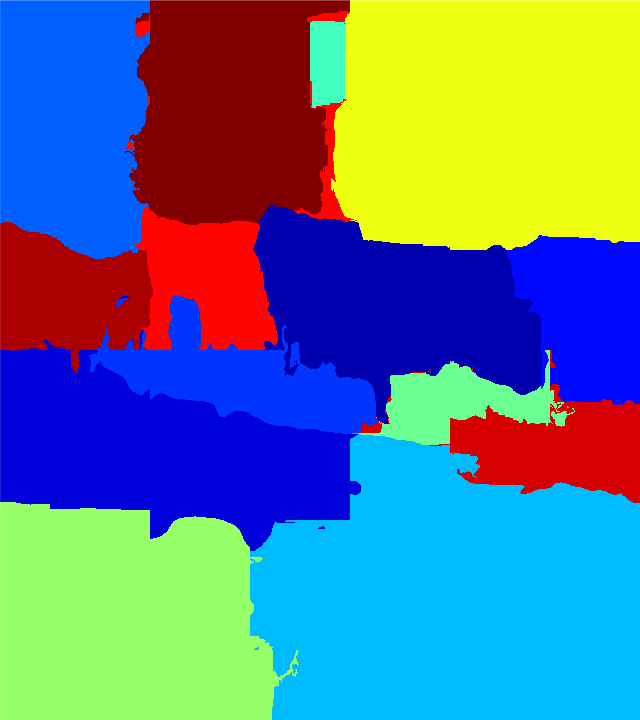} &
\includegraphics[width=0.095\linewidth]{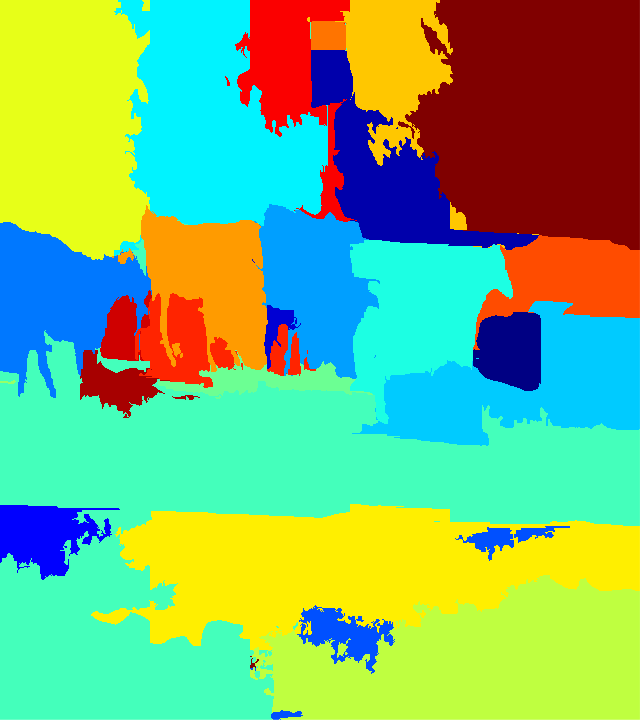}
\\
\multirow{2}{*}{\begin{tabular}[c]{@{}c@{}} \footnotesize{video} \\ \footnotesize{frames}  \end{tabular}} &
\multirow{2}{*}{\begin{tabular}[c]{@{}c@{}} \footnotesize{ground} \\ \footnotesize{truth}  \end{tabular}} &
\multirow{2}{*}{\begin{tabular}[c]{@{}c@{}} \footnotesize{GBH} \\ \cite{grundmann2010efficient} \end{tabular}} &
\multirow{2}{*}{\begin{tabular}[c]{@{}c@{}} \footnotesize{moseg} \\ \cite{ochs2011object} \end{tabular}} &
\multirow{2}{*}{\begin{tabular}[c]{@{}c@{}} \footnotesize{SAS} \\ \cite{li2012segmentation} \end{tabular}} &
\multirow{2}{*}{\begin{tabular}[c]{@{}c@{}} \footnotesize{4D-seg} \\ \cite{2014-Hickson-EHGSRV} \end{tabular}} &
\multicolumn{4}{c}{\footnotesize{proposed EASVS}} 
\\
& & & & & & \scriptsize{(BPR)} & \scriptsize{(VPR)} & \scriptsize{(AM-BVPR)} & \scriptsize{(HM-BVPR)}
\end{tabular}
\caption{Examples of the proposed EASVS optimized for different metrics compared to the state-of-the-art algorithms.
Note how GBH outlines the object boundaries but tends to over-segment, while moseg produces under-segmentations and fails to extract objects without significant motion. The static combination scheme SAS cannot strike good compromised parameters across all videos, which results in degraded results. 4D-seg~\cite{2014-Hickson-EHGSRV} is a clear leap forward but suffers from some of the drawbacks of GBH.
Our EASVS benefits the learning framework and the adaptive nature for a better output.
}
\label{fig:results}
\end{table*}

\subsection{EASVS and the state-of-the-art}\label{sec:sotacomp}

Our adaptive combination of pooled segmentation outputs poses the question of which measure to use for learning. As mentioned in Section~\ref{sec:metricsall}, the BPR and VPR measures may push for adaptive algorithms with better boundaries or temporally-consistent volumes. Averaging BPR and VPR may balance the two aspects, which we may achieve by arithmetic (AM-BVPR) or harmonic mean (HM-BVPR).

In Table~\ref{tab:vid_results}, we illustrate performance of EASVS against moseg, GBH and SAS, measured according to the four available metrics (BPR, VPR, AM-BVPR, HM-BVPR). For EASVS, the measured performance statistic has also been respectively used for learning the adaptive ensemble segmentation model. (Since our approach involves learning, we address train/test splits with three-fold cross validation and the results are averaged on three folds.) The results in the table match the intuition that only an adaptive combination can successfully address all videos. Furthermore, our proposed EASVS outperforms a recent depth video segmentation method~\cite{2014-Hickson-EHGSRV} (4D-seg) by more than 50\% on all measures, as well as a recent video co-segmentation algorithm \cite{chiu2013coseg} that we run on each video stereo pair by 65\%. This is confirmed by the qualitative examples shown in Table~\ref{fig:results}.

We delve further into the understanding of the potential result improvements within the EASVS framework with an oracle. In more details, we allow our algorithm to estimate the optimal segmentation-pool combination-parameters ($\alpha$ and $\beta$) by accessing the ground truth performance measure $P$ for each stereo video sequence. The higher oracle performance by up to 70\% (with the current representation and quadratic regressors) anticipate future improvements with richer models and more data.

\subsection{Deeper analysis of EASVS}\label{sec:indepth}

In Table~\ref{tab:adapt_results}, we provide additional insights into EASVS. First, we experiment with 1) no depth and 2) fixed depth contribution. The performance drops by $11.5\%$ for 1) and $4.2\%$ for 2) in HM-BVPR. This shows the importance of the depth cue within the full system, which benefits for videos with motion or appearance ambiguities. Additionally, this speaks in favor a the adaptive strategy. (Cf.\ 4D-seg \cite{2014-Hickson-EHGSRV} also leverages depth but cannot reach the same performance as the adaptive depth combination.)

Second, we fix the combination parameters 3) $\alpha$, 4) $\beta$, and 5) both $\alpha, \beta$ to the single best values determined on the training set, therefore limiting the system adaptivity. The performance drops by $6.3\%$, $4.2\%$, and $5.6\%$ respectively. (Please note that although conceptually fixing both $\alpha$ and $\beta$ is the same as the SAS baseline, but for SAS we use default parameters in the code of \cite{li2012segmentation}.) Once again, we find that adaptivity is therefore crucial for the performance of our system and that both adaptive aspects are strictly needed: weighting the pooled segmentation ($\alpha$) and measuring similarity of the resulting superpixels ($\beta$).
\section{Conclusions}

We have considered the emerging topic of consumer stereo cameras and proposed a benchmark to evaluate progress for the task of segmentation with this interesting type of data. The dataset is challenging and it includes diverse visual cues and camera setups. None of the existing segmentation algorithms can perform well in all conditions.

Furthermore, we have introduced a novel efficient and adaptive stereo video segmentation algorithm. Our method is capable of combining optimally a pool of segmentation outputs from a number of "expert" algorithms. The quality of results highlights that combining single algorithms is promising and that research on such a framework is perfectly orthogonal to pushing performance in the single niches, e.g.\ motion segmentation, image segmentation, supervoxelization etc.

\bibliographystyle{splncs}
\bibliography{egbib}

\end{document}